\documentclass[11pt,a4paper]{article}
\usepackage[hyperref]{emnlp2020}
\usepackage{times}
\usepackage{latexsym}
\usepackage{amsmath, amsfonts, amssymb}
\usepackage{graphicx}
\usepackage{multirow}
\usepackage{subcaption}

\usepackage{microtype}

\aclfinalcopy 

\newcommand{\softmax}{{\mathit{softmax}}}

\newcommand{\unidir}[2]{{#1$\rightarrow$#2}}
\newcommand{\OT}{{T_{\mathit{opt}}}}
\renewcommand{\dag}{{$^{\dagger}$}}

\newcommand{\Sec}[1]{{Section~\ref{sec:#1}}}
\newcommand{\Tab}[1]{{Table~\ref{tab:#1}}}
\newcommand{\Fig}[1]{{Figure~\ref{fig:#1}}}
\newcommand{\Eq}[1]{{Equation~(\ref{eq:#1})}}

\title{Softmax Tempering for Training Neural Machine Translation Models}

\author{Raj Dabre \qquad Atsushi Fujita\\
National Institute of Information and Communications Technology \\
3-5 Hikaridai, Seika-cho, Soraku-gun, Kyoto, 619-0289, Japan\\
\tt{firstname.lastname@nict.go.jp}}

\date{}

\begin{document}
\maketitle
\begin{abstract}
Neural machine translation (NMT) models are typically trained using a softmax cross-entropy loss where the softmax distribution is compared against smoothed gold labels. In low-resource scenarios, NMT models tend to over-fit because the softmax distribution quickly approaches the gold label distribution. To address this issue, we propose to divide the logits by a temperature coefficient, prior to applying softmax, during training. In our experiments on 11 language pairs in the Asian Language Treebank dataset and the WMT 2019 English-to-German translation task, we observed significant improvements in translation quality by up to 3.9 BLEU points. Furthermore, softmax tempering makes the greedy search to be as good as beam search decoding in terms of translation quality, enabling 1.5 to 3.5 times speed-up. We also study the impact of softmax tempering on multilingual NMT and recurrently stacked NMT, both of which aim to reduce the NMT model size by parameter sharing thereby verifying the utility of temperature in developing compact NMT models. Finally, an analysis of softmax entropies and gradients reveal the impact of our method on the internal behavior of NMT models.
\end{abstract}

\section{Introduction}
\label{sec:intro}

Neural machine translation (NMT) \citep{DBLP:journals/corr/SutskeverVL14,DBLP:journals/corr/BahdanauCB14:original} enables end-to-end training of translation models and is known to give state-of-the-art results for a large variety of language pairs. NMT for high-resource language pairs is straightforward: choose an NMT architecture and implementation, and train a model on all existing data. In contrast, for low-resource language pairs, this does not work well due to the inability of neural networks to generalize from small amounts of data. One reason for this is the strong over-fitting potential of neural models \citep{DBLP:conf/emnlp/ZophYMK16:original,koehn-knowles-2017-six}.

There are several solutions that address this issue of which the two most effective ones are transfer learning and model regularization. Transfer learning can sometimes be considered as data regularization and comes in the form of monolingual or cross-lingual (multilingual) transfer learning \citep{DBLP:conf/emnlp/ZophYMK16:original, DBLP:conf/icml/SongTQLL19}, pseudo-parallel data generation (back-translation) \citep{sennrich-haddow-birch:2016:P16-11}, or multi-task learning \citep{eriguchi-etal-2017-learning}. On the other hand, model regularization techniques place constraints on the learning of model parameters in order to aid the model to learn robust representations that positively impact model performance. Among existing model regularization methods, dropout \citep{10.5555/2627435.2670313} is most commonly used and is known to be effective regardless of the size of data. We thus focus on designing a technique that can complement dropout especially in an extremely low-resource situation.

The most common way to train NMT models is to minimize a softmax cross-entropy loss, i.e., cross-entropy between the softmax distribution and the smoothed label distribution typically represented with a one-hot vector. In other words, the NMT model is trained to produce a softmax distribution that is similar to the label. In high-resource settings, this may never happen due to the diversity of label sequences.  However, in low-resource settings, due to lack of the diversity, there is a high chance of this occurring and over-fitting is said to take place. We consider that a simple manipulation of the softmax distribution may help prevent it.

This paper presents our investigation into \emph{softmax tempering} \citep{DBLP:journals/corr/HintonVD15} during training NMT models in order to address the over-fitting issue. Softmax tempering is realized by simply dividing the pre-softmax logits with a positive real number greater than 1.0.  This leads to a smoother softmax probability distribution, which is then used to compute the cross-entropy loss. Softmax tempering has been devised and used regularly in knowledge distillation \citep{DBLP:journals/corr/HintonVD15,kim-rush-2016-sequence}, albeit for different purposes. We regard softmax tempering as a means of deliberately making the softmax distribution noisy during training with the expectation that this will have a positive impact on the final translation quality.

We primarily evaluate the utility of softmax tempering on extremely low-resource settings involving English and 11 languages in the Asian Languages Treebank (ALT) \citep{ALT:16}. Our experiments reveal that softmax tempering with a reasonably high temperature improves the translation quality. Furthermore, it makes the greedy search performance of the models trained with softmax tempering comparable to or better than the performance of the beam search using the models that are trained without softmax tempering, enabling faster decoding.

We then expand the scope of our study to high-resource settings, taking the WMT 2019 English-to-German translation task, as well as multilingual settings using the ALT data.
We also show that softmax tempering improves the performance of NMT models using recurrently stacked layers that heavily share parameters.
Furthermore, we clarify the relationship between softmax tempering and dropout, i.e., the most widely used and effective regularization mechanism.
Finally, we analyze the impact of softmax tempering  on the softmax distributions and on the gradient flows during training.

\section{Related Work}
\label{sec:relwork}
The method presented in this paper is a training technique aimed to improve the quality of NMT models with a special focus on the performance of low-resource scenarios.

Work on knowledge distillation \citep{DBLP:journals/corr/HintonVD15} for training compact models is highly related to our application of softmax tempering. However, the purpose of softmax tempering for knowledge distillation is to smooth the student and teacher distributions which is known to have a positive impact on the quality of student models. In our case, we use softmax tempering to make softmax distributions noisy during training a model from scratch to avoid over-fitting. In the context of NMT, \citet{kim-rush-2016-sequence} did experiment with softmax tempering.  However, their focus was on model compression and they did not experiment with low-resource settings. In contrast, our application of softmax tempering does have a strong positive impact on decoding speed because we observed that greedy search performs as well as beam search, similarly to \citet{kim-rush-2016-sequence}. We refer readers to orthogonal methods for speeding up NMT, such as weight pruning \citep{see-etal-2016-compression}, quantization \citep{Lin:2016:FPQ:3045390.3045690}, and binarization \citep{Courbariaux:16, oda-etal-2017-neural}.

We regard softmax tempering as a regularization technique, since it adds noise to the NMT model. Thus, it is related to techniques, such as $L_{N}$ regularization \cite{10.1145/1015330.1015435}, dropout \cite{10.5555/2627435.2670313}, and tuneout \cite{miceli-barone-etal-2017-regularization}. The most important aspect of our method is that it is only applied at the softmax layer whereas other regularization techniques add noise to several parts of the entire model. Furthermore, our method is intented to complement the most popular technique, i.e., dropout, and not necessarily replace it.

We primarily focus on low-resource language pairs and in this context data augmentation via back-translating additional monolingual data \citep{sennrich-haddow-birch:2016:P16-11} to generate pseudo-parallel corpora is one of the most effective approaches.  In contrast, our method does not need any additional data on top of the given parallel data. In the literature, multilingualism has been successfully leveraged for improving translation quality \citep{DBLP:journals/corr/FiratCB16:original,DBLP:conf/emnlp/ZophYMK16:original,dabre-etal-2019-exploiting}, and our method could possibly complement the impact of multilingualism due to its data and language independent nature. Recently, pre-training on monolingual data \citep{devlin-etal-2019-bert,DBLP:conf/icml/SongTQLL19,mao-etal-2020-jass} has been proven to strongly improve the performance of extremely low-resource language pairs. However, this requires enormous time and resources to pre-train large models. In contrast, our method can help improve performance if one does not possess the resources to perform massive pre-training, even though it could also be used on top of pre-training based methods.

\section{Softmax Tempering}
\label{sec:method}

Softmax tempering \citep{DBLP:journals/corr/HintonVD15} consists of two tiny changes in the implementation of the training phase of any neural model used for classification.

Assume that $D_{i}\in\mathbb{R}^{v}$ is the output of the decoder for the $i$-th word in the target language sentence, $Y_{i}$, where $v$ stands for the target vocabulary size, and that $P_{i} = P(Y_{i}|Y_{<i},X)=\softmax(D_{i})$ represents the softmax function producing the probability distribution, where $X$ and $Y_{<i}$ indicate the given source sentence and the past decoder output, respectively.
Let $L_{i} \in \mathbb{R}^{v}$ be the one-hot reference label for the $i$-th prediction. Then, the cross-entropy loss for the prediction is computed as $\mathit{loss}_{i} = - \langle \log(P_{i}), L_{i} \rangle$, where $\langle \cdot , \cdot \rangle $ is the inner product of two vectors.

Let $T\in\mathbb{R}^{+}$ be the temperature hyper-parameter. Then, the prediction with softmax tempering ($P^{\mathit{temp}}_{i} = P^{\mathit{temp}}(Y_{i}|Y_{<i},X)$) and the corresponding cross-entropy loss ($\mathit{loss}_{i}^{\mathit{temp}}$) are formalized as follows.
\begin{align}
P^{\mathit{temp}}_{i} &= \softmax(D_{i}/T), \label{eq:prediction}\\
\mathit{loss}_{i}^{\mathit{temp}} &= -\langle \log(P^{\mathit{temp}}_{i}), L_{i} \rangle \cdot T \label{eq:loss}
\end{align}

By referring to \Eq{prediction}, when $T$ is greater than 1.0, the logits, $D_{i}$, are down-scaled which leads to a smoother probability distribution before loss is computed. The smoother the distribution becomes, the higher its entropy is and hence the more uncertain the prediction is. Because loss is to be minimized, back-propagation will force the model to generate logits to counter the smoothing effect of temperature. During decoding with a model trained in this way, the temperature coefficient is not used and the logits will be such that they yield a sharper softmax distribution compared to those of a model trained without softmax tempering.

The gradients are altered by tempering, and we thus re-scale the loss by the temperature as shown in \Eq{loss}. This is inspired by the loss scaling method used in knowledge distillation \citep{DBLP:journals/corr/HintonVD15}, where both the student and teacher's softmaxes are tempered and the loss is multiplied by the square of the temperature.

\section{Experiment}
\label{sec:exp}

To evaluate the effectiveness of softmax tempering, we conducted experiments on both low-resource and high-resource settings.

\subsection{Datasets}
\label{sec:datasets}
We experimented with the Asian Languages Treebank (ALT),\footnote{\url{http://www2.nict.go.jp/astrec-att/member/mutiyama/ALT/ALT-Parallel-Corpus-20190531.zip}} comprising English (En) news articles consisting of 18,088 training, 1,000 development, and 1,018 test sentences manually translated into 11 Asian languages: Bengali (Bn), Filipino (Fil), Indonesian (Id), Japanese (Ja), Khmer (Km), Lao (Lo), Malay (Ms), Burmese (My), Thai (Th), Vietnamese (Vi), and Chinese (Zh). We focused on translation to and from English to each of these 11 languages.  As a high-resource setting, we also experimented with the WMT 2019 English-to-German (\unidir{En}{De}) translation task.\footnote{\url{http://www.statmt.org/wmt19/translation-task.html}} For training, we used the Europarl and the ParaCrawl corpora containing 1.8M and 37M sentence pairs, respectively. For evaluation, we used the WMT 2019 development and test sets consisting of 2,998 and 1,997 lines, respectively.

\subsection{Implementation Details}
\label{sec:implementation}
We evaluated softmax tempering on top of the Transformer model \citep{NIPS2017_7181}, because it gives the state-of-the-art results for NMT.
More specifically, we employed the following models.
\begin{itemize}\itemsep=0mm
\item \unidir{En}{XX} and \unidir{XX}{En} ``Transformer Base'' models where XX is an Asian language in the ALT dataset.
\item \unidir{En}{De} ``Transformer Base'' and ``Transformer Big'' models.
\end{itemize}

We modified the code of the Transformer model in the tensor2tensor v1.14.  For ``Transformer Base'' and ``Transformer Big'' models, we used the hyper-parameter settings in \emph{transformer\_base\_single\_gpu} and \emph{transformer\_big\_single\_gpu}, respectively where \textbf{label smoothing of 0.1 is used by default}. We used the internal sub-word tokenization mechanism of tensor2tensor with separate source and target language vocabularies of size 8,192 and 32,768 for low-resource and high-resource settings, respectively.
We trained our models for each of the softmax temperature values, 1.0 (default softmax), 1.2, 1.4, 1.6, 1.8, 2.0, 3.0, 4.0, 5.0, and 10.0. We used early-stopping on the BLEU score \citep{Papineni:2002:BMA:1073083.1073135} for the development set which was evaluated every after 1k iterations. Our early-stopping mechanism halts training when the BLEU score does not vary by over 0.1 BLEU over 10 consecutive evaluation steps.
For decoding, we averaged the final 10 checkpoints, and evaluated beam search with beam sizes (2, 4, 6 ,8, 10, and 12) and length penalties (0.6, 0.7, 0.8, 0.9, 1.0, 1.1, 1.2, 1.3, and 1.4) and greedy search.

\begin{table*}[t]
\centering
\scalebox{.75}{
\begin{tabular}{ll|rrrrrrrrrrr}
\hline
\multirow{2}{*}{$T$} &\multirow{2}{*}{Decoding} & \multicolumn{11}{c}{En-to-XX} \\
& & \multicolumn{1}{c}{Bn} & \multicolumn{1}{c}{Fil} & \multicolumn{1}{c}{Id} & \multicolumn{1}{c}{Ja} & \multicolumn{1}{c}{Km} & \multicolumn{1}{c}{Lo} & \multicolumn{1}{c}{Ms} & \multicolumn{1}{c}{My} & \multicolumn{1}{c}{Th} & \multicolumn{1}{c}{Vi} & \multicolumn{1}{c}{Zh} \\\hline
1.0   &Greedy & 3.5 & 24.3 & 27.4 & 13.4 & 19.3 & 11.5 & 31.5 & 8.3 & 13.7 & 24.0 & 10.4 \\
1.0   &Beam   & 4.2 & \textbf{25.9} & 28.7 & 15.1 & \textbf{21.5} & \textbf{13.1} & 32.8 & 9.1 & \textbf{16.0} & \textbf{26.6} & 12.1 \\
$\OT$ &Greedy & 4.5 & 25.7 & 29.5\dag & 15.5 & 20.7 & 12.2 & 33.7\dag & 9.3 & 15.6 & 25.7 & 12.8\dag \\
$\OT$ &Beam   & \textbf{4.6}\dag & \textbf{25.9} & \textbf{29.6}\dag & \textbf{15.9}\dag & 21.2 & 12.2 & \textbf{33.8}\dag & \textbf{9.7} & 15.8 & 26.1 & \textbf{13.0}\dag \\
\multicolumn{2}{l|}{Value for $\OT$} & 5.0 & 3.0 & 4.0 & 4.0 & 3.0 & 10.0 & 4.0 & 5.0 & 5.0 & 5.0 & 4.0 \\
\hline\hline
\multirow{2}{*}{$T$} &\multirow{2}{*}{Decoding} & \multicolumn{11}{c}{XX-to-En} \\
& & \multicolumn{1}{c}{Bn} & \multicolumn{1}{c}{Fil} & \multicolumn{1}{c}{Id} & \multicolumn{1}{c}{Ja} & \multicolumn{1}{c}{Km} & \multicolumn{1}{c}{Lo} & \multicolumn{1}{c}{Ms} & \multicolumn{1}{c}{My} & \multicolumn{1}{c}{Th} & \multicolumn{1}{c}{Vi} & \multicolumn{1}{c}{Zh} \\\hline
1.0   &Greedy & 7.1 & 22.2 & 25.1 & 8.7  & 14.9 & 9.8  & 27.4 & 7.8  & 10.5 & 19.4 & 9.4  \\
1.0   &Beam   & 8.5 & 24.1 & 26.5 & 10.1 & 16.5 & \textbf{11.9} & 28.6 & 9.4  & 12.5 & 21.1 & 11.0 \\
$\OT$ &Greedy & 9.1 & 24.7 & 27.5\dag & 11.0\dag & 16.8 & 11.4 & 29.7\dag & 11.7\dag & 12.2 & 21.3 & 11.5 \\
$\OT$ &Beam   & \textbf{9.3}\dag& \textbf{25.0}\dag & \textbf{27.6}\dag & \textbf{11.3}\dag & \textbf{17.1} & 11.7 & \textbf{29.9}\dag & \textbf{12.0}\dag & \textbf{12.6} & \textbf{21.5} & \textbf{12.1}\dag \\
\multicolumn{2}{l|}{Value for $\OT$} & 5.0 & 4.0 & 3.0 & 5.0 & 4.0 & 5.0 & 4.0 & 4.0 & 4.0 & 4.0 & 4.0 \\
\hline
\end{tabular}%
}
\caption{BLEU scores obtained by non-tempered ($T=1.0$) and tempered ($T=\OT>1.0$) NMT models with greedy and beam search for the ALT \unidir{En}{XX} and \unidir{XX}{En} translation tasks, where XX is one of the Asian languages in the ALT dataset. Best BLEU scores are in bold. ``{\dag}'' marks scores that are significantly ($p\le 0.05$) better than non-tempered model's (oracle) beam search scores.}
\label{tab:alt-results-low}
\end{table*}

\subsection{Evaluation Criteria}

We evaluated translation quality of each model using BLEU \citep{Papineni:2002:BMA:1073083.1073135} provided by \emph{SacreBleu} \cite{post-2018-call}.
The optimal temperature ($\OT$) for the tempered model was determined based on greedy search BLEU score on the development set.
On the other hand, two hyper-parameters for beam search, i.e., beam size and length penalty, are not searched.
The best beam search results reported in this paper are determined after computing BLEU scores on the test set, i.e., oracle.
We used statistical significance testing\footnote{\url{https://github.com/moses-smt/mosesdecoder/blob/master/scripts/analysis/bootstrap-hypothesis-difference-significance.pl}} to determine if differences in BLEU are significant.

\subsection{Results in Low-Resource Settings}
\label{sec:lrresults}
\Tab{alt-results-low} shows the greedy and beam search BLEU scores along with the optimal temperature ($\OT$) for translation to and from Asian languages and compare them against those obtained by non-tempered models.  Except for a few language pairs, the greedy search BLEU scores of the best performing tempered models are higher than the beam search BLEU scores of non-tempered models.

\begin{figure}[t]
    \centering
    \includegraphics[width=.95\columnwidth]{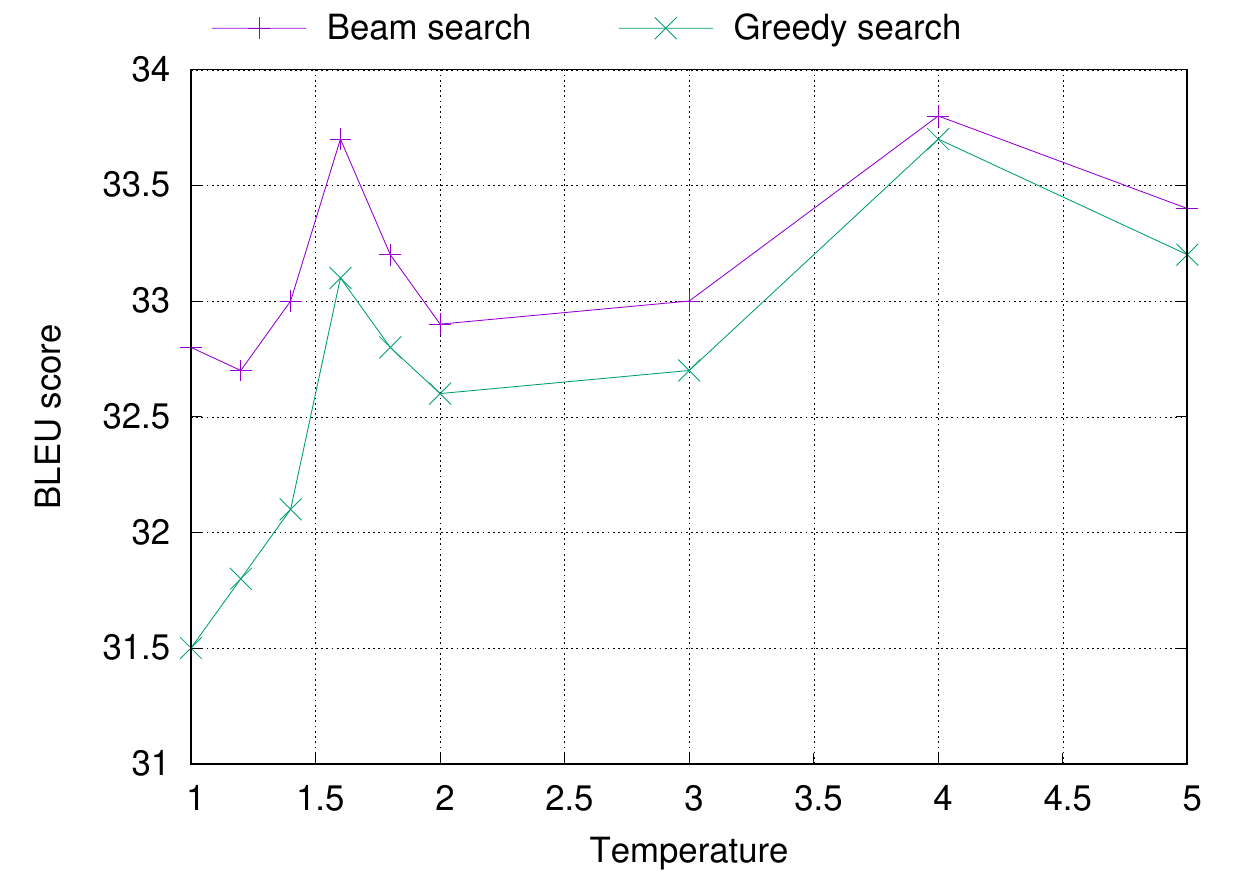}
    \caption{Greedy and beam search BLEU scores with temperature for \unidir{En}{Ms} translation.}
    \label{fig:enmsvariation}
\end{figure}

\Fig{enmsvariation} shows how the greedy and beam search results vary with the temperature, taking \unidir{En}{Ms} translation as an example.\footnote{We show only one translation direction due to lack of space. Kindly, refer to our supplementary material for all translation directions.}  As the temperature is raised, the greedy search BLEU score increases and begins approaching the beam search score. At temperature values between 2.0 and 5.0, not only does the greedy search score exceed that of a non-tempered model, but that the greedy and beam search scores are barely any different. However, increasing the temperature beyond 10.0 always has a negative effect on the translation quality, because it leads to a highly smoothed probability distribution, quantified by high entropy, that does not seem to be useful for NMT training.

Consequently, we conclude that training with reasonably high temperature (between 2.0 and 5.0), softmax tempering has a positive impact on translation quality for extremely low-resource settings.

We also computed the similarity between the greedy and beam search (beam size 4, length penalty 1.0) translations by computing the BLEU score between them, regarding the greedy search results as references. \Tab{beamgreedybleugap} reports on the greedy-beam search similarities for 4 translation directions for several different values of temperature. For non-tempered models, the BLEU scores were around 30 to 60, and as the temperature increases, so does the BLEU score. This indicates that greedy and beam search results grow to be similar and the most likely reason is that training with softmax tempering forces the model to be extremely precise.

\begin{table}[t]
\centering
\scalebox{.75}{
\begin{tabular}{c|cccc}
\hline
$T$ &\unidir{Vi}{En} &\unidir{En}{Vi} &\unidir{Ja}{En} &\unidir{En}{Ja}\\
\hline
1.0 & 53.4 & 61.5 & 32.7 & 40.5 \\
2.0 & 72.6 & 79.2 & 53.2 & 59.6 \\
3.0 & 81.9 & 84.1 & 67.2 & 69.7 \\
5.0 & 86.7 & 89.4 & 75.4 & 80.0 \\
\hline
\end{tabular}%
}
\caption{Similarity between the beam and greedy search results on the test set, indicated by BLEU score.}
\label{tab:beamgreedybleugap}
\end{table}

\begin{table}[t]
\centering
\scalebox{.75}{
\begin{tabular}{lll|rr}
\hline
\multirow{2}{*}{Model} &\multirow{2}{*}{Training} &\multirow{2}{*}{$T$} &\multicolumn{2}{c}{BLEU}\\
&&&Greedy &Beam\\
\hline
\multirow{4}{*}{Base}
&\multirow{2}{*}{EP}
 &1.0 &23.6 &25.8\\
&&1.4 &25.5 &\textbf{27.2}\dag\\\cline{2-5}
&\multirow{2}{*}{EP+PC}
 &1.0 &28.2 &29.2\\
&&1.2 &29.1 &\textbf{30.1}\dag\\
\hline
\multirow{4}{*}{Big}
&\multirow{2}{*}{EP}
 &1.0 &26.8 &29.4\\
&&1.2 &29.1 &\textbf{30.2}\dag\\\cline{2-5}
&\multirow{2}{*}{EP+PC}
 &1.0 &32.7 &33.7\\
&&1.2 &33.6 &\textbf{34.5}\dag\\
\hline
\end{tabular}%
}
\caption{BLEU scores obtained by non-tempered ($T=1.0$) and tempered ($T=\OT>1.0$) NMT models for high-resource \unidir{En}{De} translation task trained on Europarl (EP) and ParaCrawl (PC) corpora.}
\label{tab:wmt-results-high}
\end{table}

\begin{table*}[t]
\centering
\scalebox{.75}{
\begin{tabular}{llll|ccccccccccc}
\hline
Direction &Model &$T$ &Decoding
& Bn & Fil & Id & Ja & Km & Lo & Ms & My & Th & Vi & Zh \\\hline
\multirow{8}{*}{Unidirectional}
&\multirow{4}{*}{non-RS}
 &1.0   &Greedy & 3.5 & 24.3 & 27.4 & 13.4 & 19.3 & 11.5 & 31.5 & 8.3 & 13.7 & 24.0 & 10.4 \\
&&1.0   &Beam   & 4.2 & \textbf{25.9} & 28.7 & 15.1 & \textbf{21.5} & \textbf{13.1} & 32.8 & 9.1 & \textbf{16.0} & \textbf{26.6} & 12.1 \\
&&$\OT$ &Greedy & 4.5 & 25.7 & 29.5\dag & 15.5 & 20.7 & 12.2 & 33.7\dag & 9.3 & 15.6 & 25.7 & 12.8 \\
&&$\OT$ &Beam   & \textbf{4.6} & \textbf{25.9} & \textbf{29.6}\dag & \textbf{15.9}\dag & 21.2 & 12.2 & \textbf{33.8}\dag & \textbf{9.7} & 15.8 & 26.1 & \textbf{13.0} \\
\cline{2-15}
&\multirow{4}{*}{RS}
 &1.0   &Greedy & 3.2 & 21.5 & 24.3 & 11.8 & 17.4 & 10.1 & 28.6 & 7.0 & 11.5 & 22.4 & 8.9 \\
&&1.0   &Beam   & 3.7 & \textbf{23.7} & 25.8 & \textbf{13.6} & \textbf{19.8} & \textbf{11.8} & 30.4 & 8.2 & \textbf{13.8} & \textbf{24.9} & 10.7 \\
&&$\OT$ &Greedy & 3.6 & 23.0 & 25.8 & 13.1 & 18.7 & 10.4 & 30.7 & 8.3 & 13.5 & 23.1 & 10.6 \\
&&$\OT$ &Beam   & \textbf{3.9} & 23.4 & \textbf{26.2} & \textbf{13.6} & 19.1 & 11.4 & \textbf{30.9} & \textbf{8.4} & \textbf{13.8} & 23.5 & \textbf{11.2} \\
\hline
\multirow{8}{*}{One-to-many}
&\multirow{4}{*}{non-RS}
 &1.0   &Greedy & 6.2 & 24.9 & 27.0 & 18.9 & 23.0 & 14.7 & 30.9 & 12.9 & 19.0 & 27.4 & 14.3 \\
&&1.0   &Beam   & 7.1 & 26.5 & 28.6 & 21.1 & 24.7 & 15.5 & 32.1 & 13.8 & 20.6 & 29.5 & 16.2 \\
&&$\OT$ &Greedy & 7.0 & 27.6\dag & 30.0\dag & 21.2 & 24.4 & 15.5 & 34.2\dag & 14.2 & 20.5 & 30.1 & 16.3 \\
&&$\OT$ &Beam   & \textbf{7.5} & \textbf{28.6}\dag & \textbf{30.3}\dag & \textbf{22.7}\dag & \textbf{25.4} & \textbf{16.2}\dag & \textbf{34.9}\dag & \textbf{15.1}\dag & \textbf{21.9}\dag & \textbf{31.6}\dag & \textbf{17.5}\dag \\
\cline{2-15}
&\multirow{4}{*}{RS}
 &1.0   &Greedy & 6.6 & 26.4 & 28.4 & 19.6 & 23.5 & 14.8 & 32.3 & 13.2 & 19.9 & 28.8 & 14.9 \\
&&1.0   &Beam   & \textbf{7.5} & \textbf{28.4} & \textbf{30.0} & \textbf{21.3} & \textbf{25.2} & 16.1 & \textbf{33.9} & 14.3 & \textbf{21.7} & \textbf{30.7} & \textbf{17.2} \\
&&$\OT$ &Greedy & 6.6 & 26.6 & 28.8 & 19.7 & 23.7 & 14.6 & 32.5 & 13.4 & 20.0 & 28.7 & 15.1 \\
&&$\OT$ &Beam   & 7.1 & 27.0 & 29.5 & 20.9 & 24.6 & \textbf{16.3} & 33.1 & \textbf{14.6} & 21.2 & 30.4 & 16.7 \\
\hline
\end{tabular}%
}
\caption{BLEU scores obtained by unidirectional and one-to-many English-to-Asian NMT models with and without recurrent stacking (RS) of layers and softmax tempering. The values for $\OT$ for multilingual models were 1.2 to 2.0, while those for unidirectional model were 3.0 to 5.0, depending on the target language.}
\label{tab:rsnmtmlnmt}
\end{table*}

\subsection{Results in High-Resource Settings}
\label{sec:rrresults}
\Tab{wmt-results-high} gives the BLEU scores for the high-resource \unidir{En}{De} translation task.
The results indicate that compared to the low-resource settings, relatively lower temperature values are effective for improving translation quality. While the improvements in translation quality are not as large as those in the low-resource settings, greedy and beam search improve by 0.8 to 2.3 BLEU points for temperature values around 1.2 to 1.6. We noticed that higher temperature values do bridge the gap between greedy and beam search performances.  However, since they also reduce translation quality, we do not recommend using high temperature values in high-resource settings.

For the models trained only on the Europarl corpus (EP), the greedy and beam search performances of the Transformer Base model starts approaching those of the Transformer Big model.

\subsection{Impact on Training and Decoding Speed}
Although training with softmax tempering makes it difficult for a model to over-fit the label distributions, we did not notice any impact on the training time. This indicates that the improvements are unrelated to longer training times.

With regard to decoding, in low-latency settings, we can safely use greedy search with tempered models given that it is as good as, if not better than, beam search using non-tempered models. Thus, by comparing the greedy and beam search decoding speeds, we can determine the benefits that softmax tempering brings in low-latency settings. Greedy search decoding of the \unidir{Vi}{En}\footnote{For ALT tasks, decoding times are very similar when translating into English due to it being a multi-parallel corpus.} test set requires 37.6s on average, whereas beam search with beam sizes of 4 and 10 require 56.4s and 138.2s, respectively. For non-tempered models, where beam search scores are higher than greedy search scores by over 2.0 BLEU points, the best BLEU scores are obtained using beam sizes between 4 and 10. Given the improved performance with greedy search, we can decode anywhere from 1.5 to 3.5 times faster. Subjecting softmax tempering to model compression methods, such as weight pruning, might further reduce decoding time.

\section{Analysis and Further Exploration}
We now focus on an intrinsic and extrinsic analyses of our method by studying its impact on extreme parameter sharing and multilingualism, its relationship to dropout, and finally on the internal working of the models during training.

\subsection{Impact on Parameter Sharing Models}
Why can regularization improve translation quality? One possible explanation is that the constraints imposed by regularization techniques force the model to more effectively utilize its existing capacity, indicated by the model parameters. This is especially important in low-resource settings. It is thus worth verifying the impact of softmax tempering, a regularization technique, on models with significantly reduced capacity.

In particular, we experimented with models that share parameters between the stacked layers, so-called recurrently stacked (RS) models \citep{DBLP:conf/aaai/DabreF19}, known to suffer from reduced translation quality for NMT. While we could have trained shallower models and/or models with small hidden sizes, they inevitably require experiments with combinations of hyper-parameters, i.e., number of layers and hidden size. In contrast, RS models are relatively recent and different from non-RS ones with regards to parameter sharing and thereby simplifying exploration.

Compare the first and second blocks of \Tab{rsnmtmlnmt} for non-RS models and their RS counterparts. It is clear that RS models are always weaker than non-RS models, but the greedy search with RS models are largely improved by the tempered training. However, unlike in the case of a non-RS model, the beam search quality either remains the same or degrades in most cases. The major difference between RS and non-RS models is in the number of parameters and we can safely say that RS models simply lack the capacity to strongly benefit from softmax tempering. Nevertheless, the greedy search performance of an RS model with an appropriate temperature setting becomes comparable to a non-RS model trained without softmax tempering.

RS models tend to be 50-60\% smaller than non-RS models \citep{DBLP:conf/aaai/DabreF19}. Thus, when one wants RS models for low-memory settings, we recommend to train them with softmax tempering.

\subsection{Impact on Multilingualism}
By training a multilingual model, we share parameters between multiple language pairs thereby reducing the amount of model capacity available for individual language pairs. Additionally, using multilingual data is a way to simulate a high-resource situation. While softmax tempering is not always useful in unidirectional high-resource translation settings (see \Sec{rrresults}), it may benefit simulated high-resource settings realized though multilingual models for extremely low-resource language pairs.

We trained a one-to-many NMT model \citep{DBLP:conf/acl/DongWHYW15} for English to 11 Asian languages by concatenating the training data of the ALT corpus after prepending each source sentence with an artificial token indicating the target language as in \citet{TACL1081}. The ALT corpus is 12-way parallel, and thus the English side contains the same sentences 11 times. We used a vocabulary with 8,192 sub-words\footnote{In our preliminary experiment with a 32,768 sub-word vocabulary for English, we obtained lower BLEU scores than those of unidirectional models. We realized that the reason was that increasing the number of language pairs did not increase the English side vocabulary and thus these sub-words resulted in a word-level vocabulary which negatively affects translation quality.} for English and another one with 32,768 sub-words for all target languages combined. As in \Sec{exp}, we trained Transformer Base models with different values for temperature.

Comparing the non-tempered models ($T=1.0$) in the first and third blocks of \Tab{rsnmtmlnmt}, it is clear that multilingual models already outperform corresponding unidirectional models by up to 6.0 BLEU points (\unidir{En}{Ja}), even though \unidir{En}{Id} and \unidir{En}{Ms}, which originally have the highest BLEU scores, suffer slightly. This highlights the utility of multilingualism in itself, since our training data is multi-parallel and introducing a new language pair does not introduce new contents.
With softmax tempering, temperature values between 1.2 and 2.0 brought improvements in translation quality by up to 6.8 BLEU points over the best unidirectional models (marked bold in the first block). Note that the optimal temperature values, $\OT$, are similar to those in the case of high-resource \unidir{En}{De} translation task. Although multilingualism simulates a high-resource setting, we observed that softmax tempering is more effective compared to a unidirectional modeling in the real high-resource setting. We thus recommend training multilingual models with softmax tempering, especially when the individual language pairs are low-resource.

The bottom block of \Tab{rsnmtmlnmt} show results when we push parameter sharing to its extreme limits combining both RS and multilingualism.
Comparing them with the third block, multilingual RS models without softmax tempering always outperform its non-RS counterpart by up to 1.9 BLEU points. However, multilingual RS models are negatively impacted by softmax tempering: the greedy search translation quality increases slightly (if at all), while the beam search translation quality degrades. Multilingual models already share parameters for 11 translation directions, significantly lowering the capacity per translation direction than unidirectional models. RS of layers further increases the burden on the multilingual model. Thus, temperature puts further burden on the model's learning which has a negative impact on performance.

Ultimately, multilingual tempered models give the best possible translations.  We thus recommend to train (a)~multilingual RS models without softmax tempering aiming at extreme compactness or (b)~multilingual non-RS models with softmax tempering for higher translation quality and faster decoding with greedy search.

\subsection{Dropout}

Softmax tempering is a kind of regularization, since it makes the model predictions noisy during training, aiming to enable better learning. We have so far experimented with softmax tempering in combination with dropout. To this end, we experimented with both low-resource and high-resource settings using softmax tempering with and without dropout to explicitly examine their complementarity.

For low-resource settings, we randomly chose \unidir{Bn}{En} and \unidir{Vi}{En} translation directions, and additionally trained unidirectional Transformer Base models using softmax tempering in various temperature settings but without dropout.  We also trained Transformer Base and Big models for the \unidir{En}{De} task without dropout.
To be precise, to disable dropout, we set attention, embedding, and layer dropouts in the model to zero during training.

The results are shown in \Tab{dropoutalt}. It is known that disabling dropout significantly deteriorates the translation quality.  However, in our \unidir{Vi}{En} and \unidir{Bn}{En} settings, higher temperature values are able to compensate for the drops. This shows that dropout and softmax tempering are complementary and that the latter acts as a regularizer similarly to the former. Note that softmax tempering alters the softmax layer, whereas dropout is applied throughout the model. This can explain why dropout has a larger impact than softmax tempering.

In the high-resource setting, in contrast, dropout does not positively impact on the translation quality. In the context of high-resource NMT, the impact of dropout on the Transformer was never explicitly investigated in detail. In such a situation, the large quantities of data might enable enough regularization. However, in the case of Transformer Base, temperature values of 1.2 to 1.6 tend to improve the greedy search performance when dropout is not used, bridging the performance gap between the Transformer Base and Transformer Big models. This shows that softmax tempering can be an alternative to dropout depending on the setting. This warrants further exploration of the relationship between data regularization and model regularization.

\begin{table}[t]
\centering
\scalebox{.75}{
\begin{tabular}{lll|c@{~~}c@{~~}cc}
\hline
\multirow{2}{*}{Dropout} & \multirow{2}{*}{$T$} &\multirow{2}{*}{Decoding}
&\multirow{2}{*}{\unidir{Bn}{En}} &\multirow{2}{*}{\unidir{Vi}{En}}
&\multicolumn{2}{c}{\unidir{En}{De}}\\
&&&&&(Base) &(Big)\\
\hline
\multirow{4}{*}{Yes}
&1.0   & Greedy & 7.1 & 19.4 & 28.2 & 32.7 \\
&1.0   & Beam   & 8.5 & 21.1 & 29.2 & 33.7 \\
&$\OT$ & Greedy & 9.1 & 21.3 & 29.1 & 33.6 \\
&$\OT$ & Beam   & \textbf{9.3}\dag & \textbf{21.5} & \textbf{30.1}\dag & \textbf{34.5}\dag \\
\hline
\multirow{4}{*}{No}
&1.0   & Greedy & 2.6 & 17.4 & 29.5 & 33.3 \\
&1.0   & Beam   & 3.1 & 19.9 & 30.7 & \textbf{34.5} \\
&$\OT$ & Greedy & \textbf{3.5} & 20.9\dag & 30.6 & 33.3 \\
&$\OT$ & Beam   & \textbf{3.5} & \textbf{21.1}\dag & \textbf{31.7}\dag & 34.1 \\
\hline
\end{tabular}%
}
\caption{BLEU scores for a subset of translation directions in the ALT tasks and \unidir{En}{De} task with and without dropout and softmax tempering.}
\label{tab:dropoutalt}
\end{table}

\begin{figure*}[t]
    \centering
    \begin{subfigure}[t]{\columnwidth}
        \centering
        \includegraphics[width=.95\columnwidth]{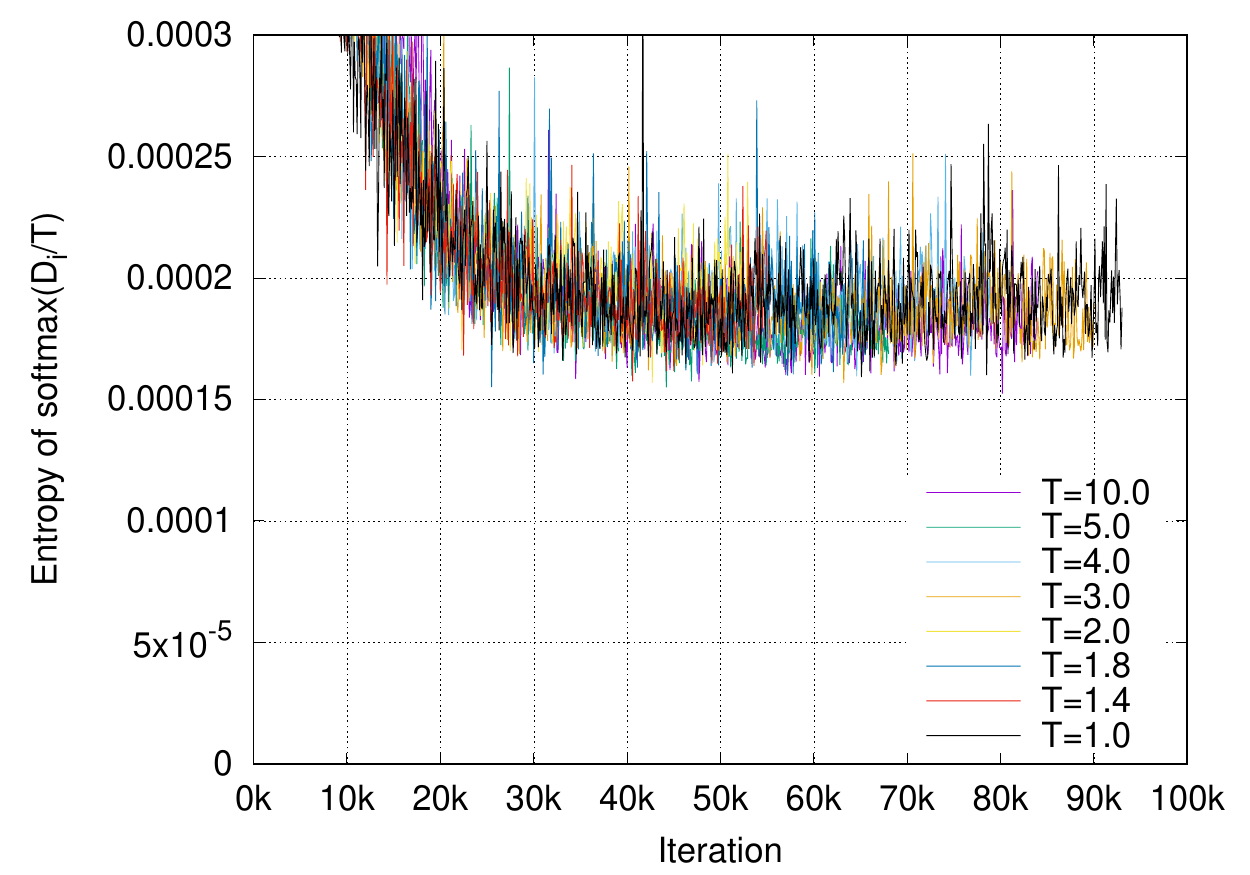}
        \label{fig:entropydistr1}
    \end{subfigure}%
    \hfill
    \begin{subfigure}[t]{\columnwidth}
        \centering
        \includegraphics[width=.95\columnwidth]{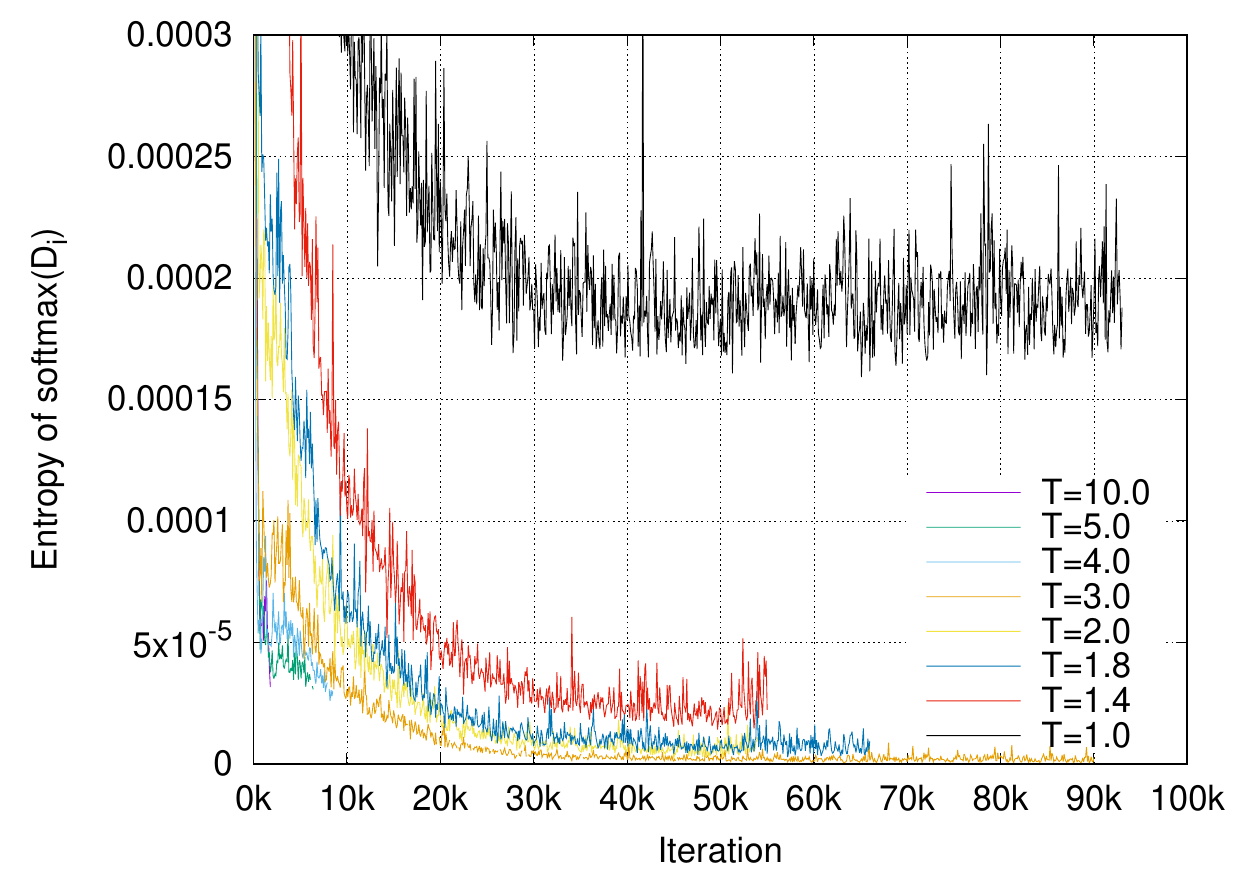}
        \label{fig:entropydistr2}
    \end{subfigure}%
    \caption{Variation of entropy: the left-hand side shows $\softmax(D_{i}/T)$ in \Eq{prediction} actually used for computing the loss during training, whereas the right-hand side shows $\softmax(D_{i})$ drawn for this analysis.}
    \label{fig:entropydistr}
\end{figure*}

\begin{figure}[t]
    \centering
    \includegraphics[width=.95\columnwidth]{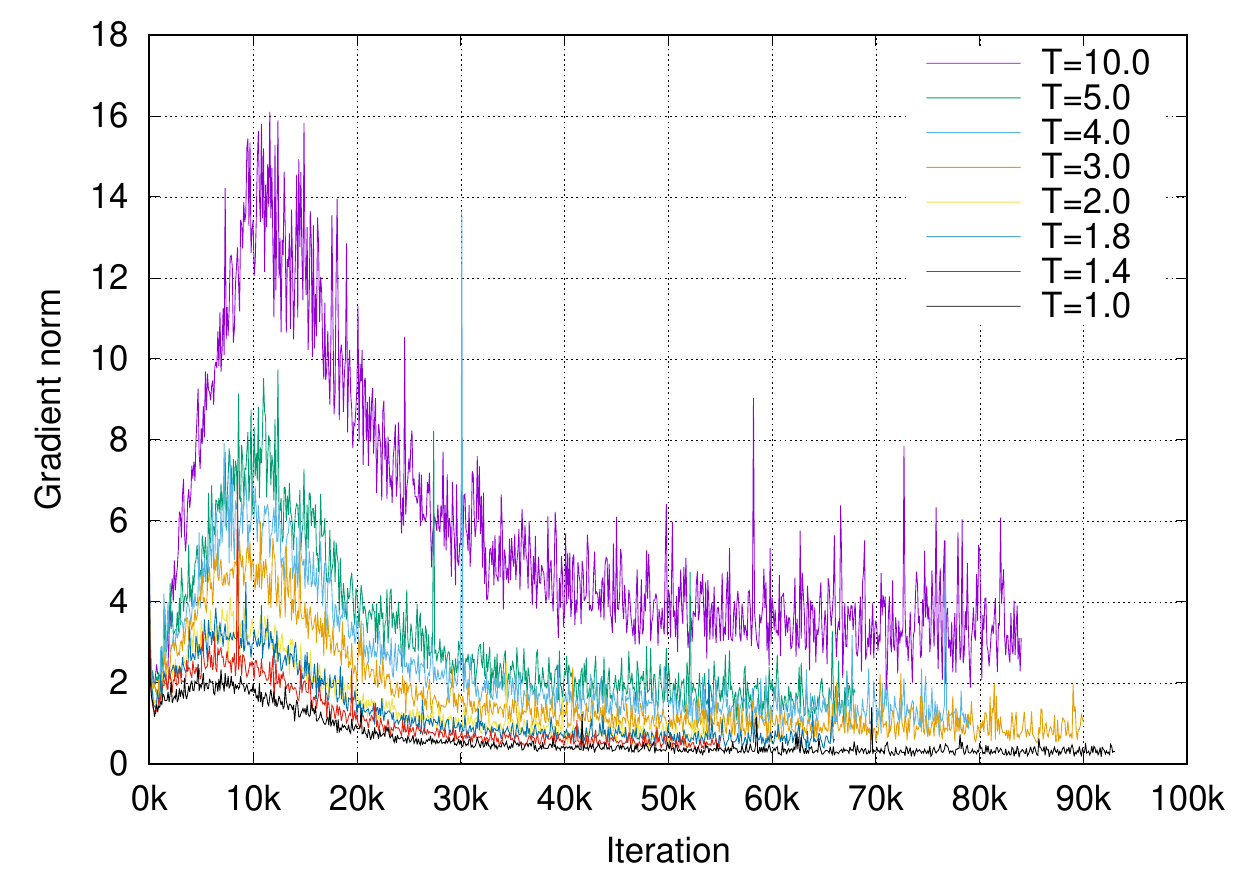}
    \caption{Global gradient norms during training models with softmax tempering.
    }
    \label{fig:graddistr}
\end{figure}

\subsection{Temperature and Model Learning}
\label{sec:internal}

We expected that tempering leads to a smoother softmax and loss minimization using such a softmax makes it sharper as training progresses. With softmax tempering, the model will continue to receive strong gradient updates even during later training stages due to the deliberate perturbation of the softmax distribution. Thus, we confirm whether our model truly behaves this way through visualizing the softmax entropies and gradient values.

\Fig{entropydistr} visualizes the variation of softmax entropy averaged over all tokens in a batch during training.
The left-hand side shows the entropy of tempered softmax distribution in \Eq{prediction}, where there is no visible differences between charts with different values for temperature, i.e., $T$.
Considering that the distribution is tempered with $T$, this indicates that the distribution of logits, $D_{i}$, is sharper when tempered with a higher $T$.
The right-hand side plots the entropy of softmax distribution derived from the logits without dividing them by $T$.  The lower entropy confirm that the distribution of logits is indeed sharper with higher $T$ and that division by $T$ as in \Eq{prediction} counters the effect of sharpening.
This means that the distribution of logits is forced to become sharper and thus produce exactly one word that the model believes the best to generate the rest of the sequence. Given the fact that translation quality is improved by softmax tempering, designing training methods that lead to sharper distribution might be useful.

\Fig{graddistr} shows the gradient norms during training with softmax tempering. This revealed that, similarly to ordinary non-tempered training, gradient norms in softmax tempering first increase during the warm-up phase of training and then gradually decrease. However, the major difference is that the norm values significantly decrease for the non-tempered training, whereas they are much higher for training with softmax tempering. Note that we re-scaled the loss for softmax tempering as in \Eq{loss}, which is one reason why the gradient norms are higher. Larger gradient norms indicate that strong learning signals are being back-propagated and this will continue as long as the softmax is forced to make erroneous decisions because of higher temperature values.

We can thus conclude that the noise introduced by softmax tempering and subsequent loss re-scaling strongly affect the behavior of NMT models which eventually have an impact on translation quality.

\section{Conclusion}
In this paper, we explored the utility of softmax tempering for training NMT models. Our experiments in low-resource and high-resource settings revealed that not only does softmax tempering lead to an improvement in the decoding quality but also bridges the gap between greedy and beam search performance. Consequently, we can use greedy search while achieving better translation quality than non-tempered models leading to 1.5 to 3.5 times faster decoding. We also explored the compatibility of softmax tempering with multilingualism and extreme parameter sharing, and explicitly investigated the complementarity of softmax tempering and dropout, where we show that softmax tempering can be an alternative to dropout in high-resource settings, while it is complementary to dropout in low-resource settings. Our analysis of the softmax entropies and gradients during training confirms that tempering gives precise softmaxes while enabling the model to learn with strong gradient signals even during late training stages.

In the future, we will explore the effectiveness of softmax tempering in other natural language processing tasks.

\section*{Acknowledgments}

A part of this work was conducted under the commissioned research program ``Research and Development of Advanced Multilingual Translation Technology'' in the ``R\&D Project for Information and Communications Technology (JPMI00316)'' of the Ministry of Internal Affairs and Communications (MIC), Japan.
Atsushi Fujita was partly supported by JSPS KAKENHI Grant Number 19H05660.

\bibliographystyle{acl_natbib}
\bibliography{emnlp2020}

\appendix

\section{Variation of Greedy and Beam Search Results with Temperature}

\begin{figure*}[t]
  \centering
  \begin{subfigure}[t]{.48\textwidth}
    \includegraphics[width=\columnwidth]{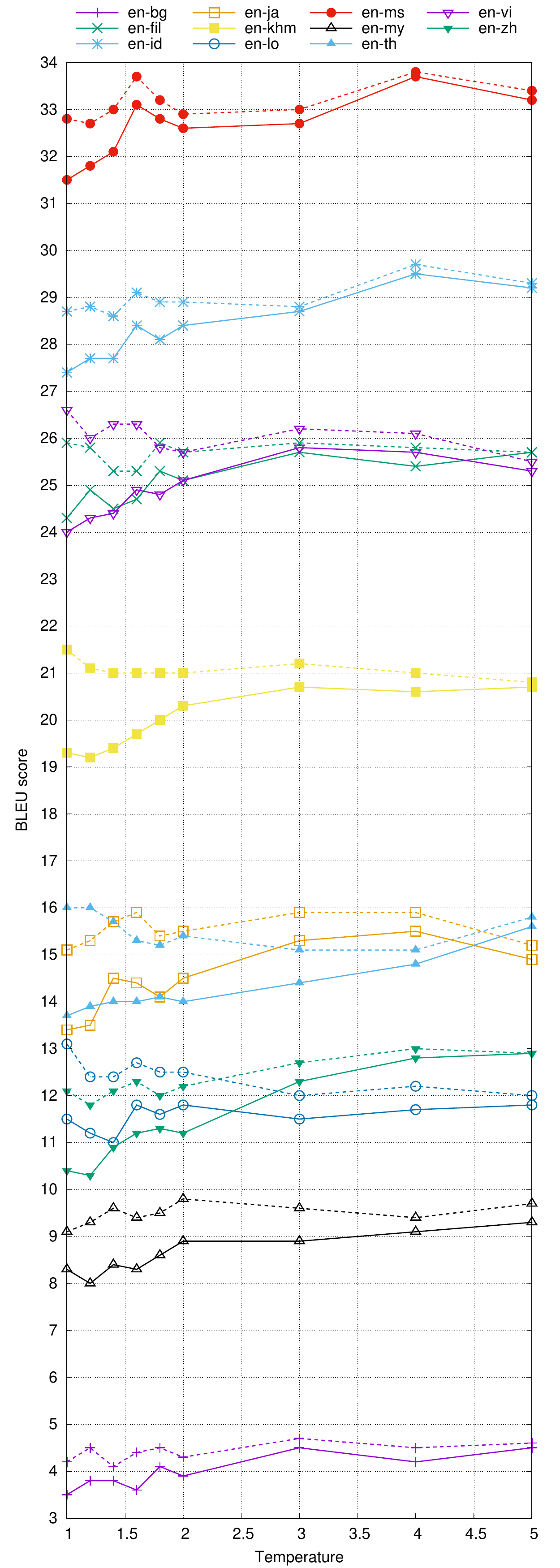}
    \caption{English-to-XX translation.}
    \label{fig:enxx}
  \end{subfigure}
  \begin{subfigure}[t]{.48\textwidth}
    \includegraphics[width=\columnwidth]{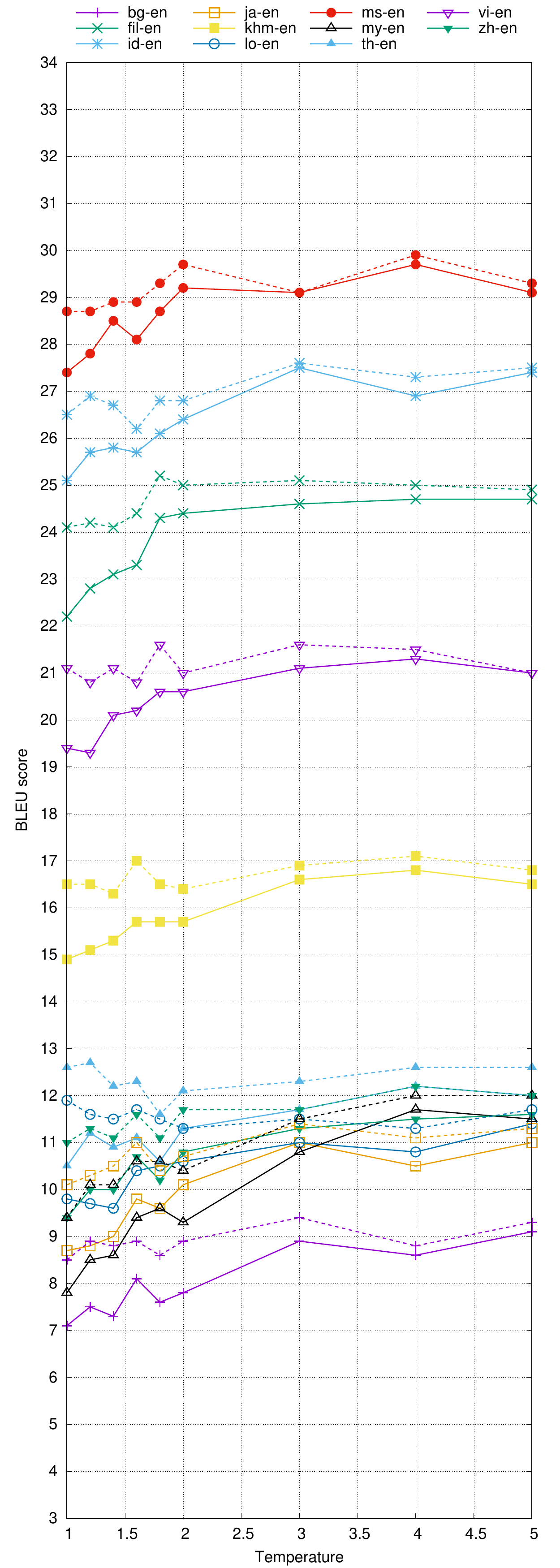}
    \caption{XX-to-English translation.}
    \label{fig:xxen}
  \end{subfigure}
  \caption{BLUE scores for all the 22 translation directions in the ALT translation tasks.  Solid and dotted lines indicate the results with greedy search and the oracle score with beam search, respectively.}
  \label{fig:ALTall}
\end{figure*}

\Fig{ALTall} shows the BLEU scores for all the 22 translation directions in the ALT translation tasks.
Solid lines in the figure show the results obtained by greedy search.
On the other hand, dotted lines are the oracle beam search result, i.e., at each temperature, the best score among 54 different combinations of beam sizes and length penalties is determined after computing BLEU score.

\end{document}